# Automated Building Heritage Assessment Using Street-Level Imagery

Kristina Dabrock[1], Tim Johansson[2,*], Anna Donarelli[3], Mikael Mangold[2,4], Noah Pflugradt[1], Jann Michael Weinand[1], Jochen Linßen[1]

[1] Forschungszentrum Jülich GmbH, Institute of Climate and Energy Systems, Jülich Systems Analysis, 52425 Jülich, Germany

[2] RISE Research Institutes of Sweden AB

[3] Uppsala University, Sweden

[4] Malmö University, Urban Studies, Sweden

* corresponding author: Tim Johansson tim.johansson@ri.se

# CRediT Author Roles

**Kristina Dabrock**: Conceptualization; Data curation; Formal analysis; Investigation; Methodology; Software; Validation; Visualization; Writing - original draft; Writing – review & editing

**Tim Johansson**: Conceptualization; Data curation; Investigation; Methodology; Software; Validation; Supervision; Writing - original draft; Writing – review & editing

**Anna Donarelli**: Conceptualization; Investigation; Validation; Visualization; Writing - original draft; Writing – review & editing

**Mikael Mangold**: Conceptualization; Investigation; Validation; Supervision; Writing – review & editing

**Noah Pflugradt**: Writing – review & editing

**Jann Michael Weinand**: Writing – review & editing

**Jochen Linßen**: Funding acquisition

# Abstract

Registration of heritage values in buildings is important to safeguard heritage values that can be lost in renovation and energy efficiency projects. However, registering heritage values is a cumbersome process. Novel artificial intelligence tools may improve efficiency in identifying heritage values in buildings compared to costly and time-consuming traditional inventories. In this study, OpenAI's large language model GPT was used to detect various aspects of cultural heritage value in facade images. Using GPT derived data and building register data, machine learning models were trained to classify multi-family and non-residential buildings in Stockholm, Sweden. Validation against a heritage expert-created inventory shows a macro F1-score of 0.71 using a combination of register data and features retrieved from GPT, and a score of 0.60 using only GPT-derived data. The methods presented can contribute to higher-quality datasets and support decision making.

# 1 Introduction

Buildings are responsible for a large share of energy usage in the European Union (EU) and also contribute significantly to greenhouse gas emissions [1]. To reach the emission reduction targets and become climate neutral by 2050, mitigation measures need to be implemented across the building stock [2]. One subgroup of buildings deserves particular attention, namely those that have a high heritage value. Energy retrofits on heritage protected buildings are often more expensive, which can lead to owners abandoning the buildings instead of renovating their buildings. Finding a balance between the protection of heritage value and energy retrofits is a challenge [3–6]. The updated EU's Energy Performance of Buildings Directive (EPBD) [7] and the Renovation Wave [8] require member states to identify heritage values in their respective building stock. Early identification of heritage values support decision making and helps avoid lost heritage values and increased costs [9,10]. However, in many countries, there is currently a lack of comprehensive registers containing heritage values [11], which makes it difficult to quantify the impacts of new regulations. For the cumbersome task of establishing such registers, an automated process for identifying buildings is preferred. Machine learning methods using street view images has been successfully applied to a variety of building-usage classifications [12–14] , and some early initiatives have been taken to set the stage for cultural-heritage classifications [14–20].

Traditionally, visual information relating to heritage values has been collected in situ by experts, examining individual buildings and their surroundings by car, cycle or on foot. An example of this approach is the inventory conducted in Halland county, Sweden (2005-2009) [21]. Ocular assessment of heritage values is only a first step, more information is necessary to add to the assessment, such as the history of the place, the significance for the community and other aspects that are not visible [22]. This process of performing surveys is costly and time-consuming. Many regions and municipalities therefore do not have updated information relating to heritage values in the building stock. Even for officially protected buildings, data is, if at all available, frequently scattered and in inhomogeneous formats. An automated approach for detecting features visible in images can help in the initial step of identifying heritage values.

The purpose of this study is to address the lack of large-scale information on heritage values by developing a method that leverages the potential of multi modal Large Language Models (LLM) demonstrated in other studies [14–17,23,24] and assessing its applicability to such a complex, partially value-based characteristic as heritage value. For this article, the following research question frames the work: Can multi modal LLMs and street view imagery be used to develop machine learning tools for predicting aspects of heritage value for automated assessments?

The case study area considered is the city of Stockholm, the capital of Sweden, where the coverage of registered heritage values is high. The intention is for the Swedish Authorities to later use the automated assessments in the EPBD implementation process. In that process expert validation will take place in workshops. In this article, validation is conducted by comparison with expert validations in the register maintained by the Stockholm City Museum on buildings heritage values.

As described in detail in Section 2, the automated assessment is done by prompting LLMs with the inclusion of a street view image and auxiliary register information about specific buildings for parameters of cultural heritage values. The parameters are analyzed using supervised conventional artificial intelligence (AI) algorithms to assign cultural heritage values useful for applications. Section 3 then presents the feature characteristics retrieved from the images using GPT as well as the classification performance. A discussion of the results is included in Section 4, while Section 5 provides the conclusions and an outlook for future analyses and potential applications.

## 1.1 Heritage value

Heritage value is defined by the Swedish National Heritage Board as a collective term for values ascribed to heritage objects, or other heritage phenomena, based on cultural-historic, social and aesthetic aspects [25]. Values should always be seen as attributed by one or more persons or institutions; they are not intrinsic in heritage objects. Heritage values are not fixed, they change as society changes, as new knowledge and views develop, and as what is considered important to preserve of history is reconsidered [22]. Eriksson [3] describes three different levels of the value concept, where ‘significance’ is used as collective concept for the combination of values connected to a place, ‘heritage values’ constitute a more functional concept which includes both tangible and intangible aspects, and ‘character-defining elements’ are material manifestations in buildings that makes it possible to understand their context and therefore adds heritage value to them. The character-defining elements are part of the assessment of heritage value and consist largely of features visible on the exterior and interior of the building. They can manifest in materials and techniques from which the building is constructed or constitute the architectural design the building expresses. Eriksson et. al. [26] developed a method where the character-defining elements were used as indicators, specifically in a decision support system for balancing energy efficiency measures with heritage values. They were identified using a template, where building elements as well as the surroundings, were described and related to their significance. They were then used as indicators for assessing vulnerability to change. Several, but not all, character-defining elements of importance for the building’s heritage value are visible in the facade and its windows, doors, ornaments, materials as well as in the roof and streetscape.

The heritage values in buildings are protected through different legislations in Sweden. The Historic Environment Act [27] protects buildings of national interest. These buildings are to be considered designated heritage buildings. Non-designated buildings, however, are existing buildings that do not qualify for protection by the Historic Environment Act. Instead, their heritage values are protected by the Planning and Building Act [28,29] through the *requirement of caution* and in some cases the *prohibition of distortion*. The requirement of caution stipulates that all changes to all existing buildings should be performed cautiously, with regard to the character of the building, maintaining technical, historical, environmental and artistic values (PBL 2010:900, c. 8 §17). It applies to all existing buildings, whether they were built 200 years ago or yesterday. In the Planning and Building Act, buildings and built environments can also be protected as *particularly valuable* from a conservation point of view if their preservation can be of a “genuine public interest” (PBL 2010:900, c. 8 § 13). If a building or a built environment is particularly valuable the prohibition of distortion applies. In the Swedish Building Regulations there is a general advice on which buildings and built environments can be defined as particularly valuable, these are buildings and built environments that:

- represent a previously common building category or construction that has now become rare
- highlight past housing conditions, social and economic conditions, working conditions and the living conditions of different groups
- illustrate past urban design and architectural ideals, values and ways of thinking
- have represented important functions or activities for the local community.

The age of the building does not in itself constitute an argument in favor of being considered particularly valuable. However, buildings constructed before the expansion of the 1920s constitute a very limited part of today's building stock. Many of these have undergone major changes. The likelihood that a reasonably well-preserved building constructed before the 1920s fulfils one of the above-mentioned criteria for a particularly valuable building is very high. Such buildings should therefore be considered particularly valuable unless it can be justified why they should not be considered as such [30]. In conclusion, the building’s age, along with its character-defining elements are of importance in an initial state of assessing a building’s heritage value. The facade provides additional information that can be useful for further assessment. Facades are characterized by different features, such as ornaments, windows, materials, doors, and portals. The characteristic features will vary depending on architectural styles and construction year [31]. For example, where on the one hand the presence of ornaments can be characteristic in one period the absence of them will be characteristic for another [31–33].

## 1.2 Machine learning

Traditional machine learning approaches require large amounts of data as well as extensive computing resources for model training [34]. The task of deriving this information, even when not limited to cities, has close connections with the domain of Urban computing. Urban computing leverages computing approaches such as neural networks and LLMs [34]. It is based on datasets from different sources and of different types, such as temporal or geographic data or images, with the goal of gaining a thorough understanding of various aspects of cities, such as energy usage [34]. Since 2015, the number of published deep learning studies in the field of urban computing have increased dramatically [34], as has the use of street-view imagery as a source of geospatial data across various domains [35], for example in analyses of urban greenery [36,37].

**Large Language Models in geospatial analysis**

The field of urban computing has expanded significantly because of the development of LLM and subsequent increased available computational power, access to massive amounts of training data and development of machine learning methods. Because LLMs are trained on a diverse amount of data, they have been successfully applied across a wide range of domains resulting in a “research paradigm shift” [34]. Especially in combination with vision capabilities, LLMs provide opportunities for novel approaches. As they are pretrained models, no training data is required, and zero-shot analyses are possible, i.e., using models directly without any additional training or fine-tuning. Even before the rise of LLMs, such as Open AI’s GPT (Generative Pre-trained Transformers), natural language processing has been employed in geospatial studies [38]. The potential of LLMs, and in particular GPT, have been the subject of numerous studies in diverse fields, including geospatial analysis [39]. Within geospatial analysis, text is the most common type of input data, followed by raster data such as images [39]. Wang et al. [39] describe that LLMs can take various roles, e.g., “reasoner” or “converter” and the recent research trend of asking LLMs to perform more complex tasks without the involvement of humans. However, studies [40] also high light some of the limitations of GPT, as it was not able to write convincing, logically correct essays about heritage value. For image analysis, GPT-4o-20240513 proved to be superior over other proprietary and open source models across most of the benchmark datasets [41].

Zeng et al. [14] point out that GPT has good overall geographic skills and Manvi et al. [38] accredit the large amount of geospatial information available in LLMs. While Roberts et al. [42], find that when asking for elevation of trajectories using latitude and longitude in the Alps, the general pattern fits, but accuracy is low and sensitive to prompting. Of relevance to the method described in this article, Manvi et al. [38] find that querying simply by coordinates has proven ineffective for retrieving information such as population, adding address information instead of using only coordinates has been shown to improve results [38]. LLMs are black-box-models, meaning that it is not possible to fully understand how they come to their result, can pose problems to the application in decision-making [39]. One approach to mitigate these issues is fine-tuning of the LLM for a specific purpose, as for example geospatial analyses [43]. Yet this requires large amounts of data, which makes this approach impractical for many applications.

**Street-view imagery for building analysis**

One specific domain at the intersection of geospatial analysis and architecture is building analysis. Duran et al. [44] report an increase in facade research using AI, including deep-learning and computer-vision techniques and conventional machine learning methods. They expect further advances through generative AI. Examples of applications using street-view data in combination with AI, such as LLMs, include urban renewable analysis [45], gentrification analysis [46], building usage [23,47], building floor number [12,23], building height [12], material [23] and construction years [13,14,23].

While the previously mentioned studies focus mainly on a single attribute. Liang et al. [23] follow the broader goal of enriching building data with various attributes, namely building usage, construction year, number of floors, and surface material. Furthermore, they do not use only one model but compare the performance of various models, including a fine-tuned vision-language model, classical convolutional neural networks, transformer-based models and a zero-shot prediction with Open AI’s GPT-4o. They find that the zero-shot prediction has a high performance and is competitive with the fine-tune models and underline the potential of using GPT, especially in cases where training data is not available.

The existing literature high light the potential of using street-view imagery in combination with AI, in particular GPT, for building analysis. There is an increasing number of studies focusing on easily quantifiable and straight-forward attributes, such as the building height or the construction year. However, more complex building characteristics such as their heritage value, is an emerging research area of its own [48,49]. Assessing the potential of using GPT for deriving complex, partially value-based building features is therefore a research gap which this study aims to address.

# 2 Methods

The approach implemented in this study leverages AI for the classification of buildings using street-level imagery. The case study is Stockholm, Sweden, and the study's scope is limited to non-residential and multifamily buildings, as they are a focus of the EPBD implementation [7]. By prompting OpenAI's GPT (gpt-4o-2024-08-06) and including images of Swedish multi-family houses and non-residential buildings from Stockholm, visual features are extracted and used to assess the heritage value of the building depicted in the images. Then, the performance of the models was evaluated by comparing the results to a test set compiled from official data sources. By analyzing the performance on this test set, the strengths and weaknesses of the proposed methods have been tested. An overview of the workflow is shown in Figure 1. The respective steps are described in detail in the following subsections.

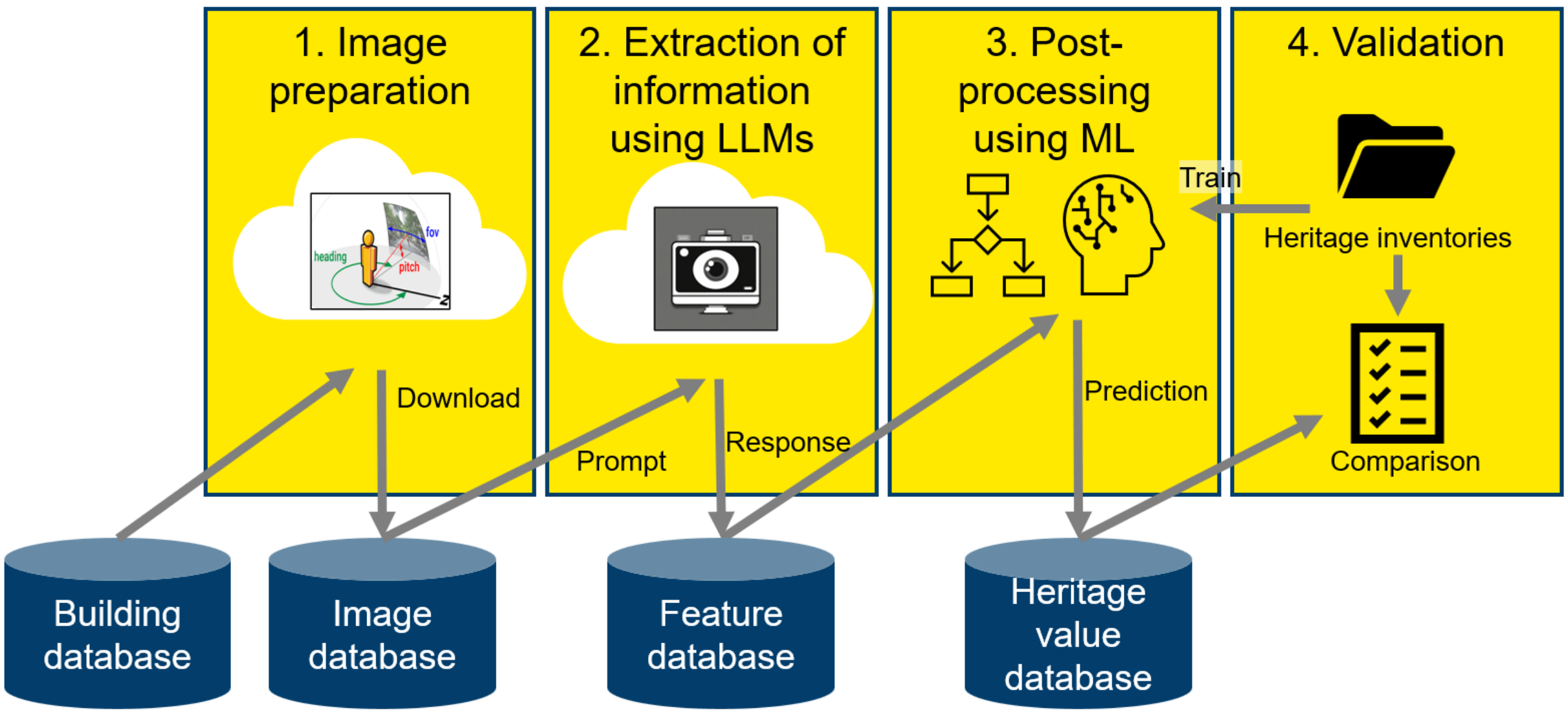

Fig. 1. Overview of the workflow for predicting heritage values for buildings.

The underlying georeferenced building dataset was created from an internal database of *RISE Research Institutes of Sweden AB*, where this study was carried out. This database includes several hundreds of building tables and geodata, including, building footprints and road lines, derived from information provided by *Boverket* [50], the Swedish National Board of Housing,

Building and Planning, and *Lantmäteriet* [51], the Swedish mapping, cadastral and land registration authority. The building database were created by using Extract, Transform and Load methods, i.e., an approach that takes data from various sources, preprocesses it, and loads it into a target database, as described in previous studies by Johansson et al. [52] and Eriksson and Johansson [53].

The building data extracted from the building database are then enriched with heritage-value information from a comprehensive, expert-based classification available for Stockholm [54]. Table 1 presents the heritage values categories, and their definition of Stockholm's classification and Figure 2 shows the distribution of the categories. The classification is based on more than visual features in the buildings' facades: the expert assessment considered the visual features in relation to historic and social contexts of the buildings. The visual features do, however, constitute an important part of the assessment, as they relate to architectural integrity and aesthetics.

The final dataset contains 15,195 entries of buildings with an assigned heritage value, 11,959 for multi-family houses and 3,236 for non-residential buildings. The building attributes include the heritage value class, height, construction year, construction year period, building category and address. 8.3% of entries fall into the blue heritage category, 42.2% into the yellow, 37.3% into the green, 3.9% into the grey, and 8.2% into the hatched category. As heritage value classifications varies between regions in Sweden, a mapping was needed to harmonize the categories for the later application of the developed methods [55], with three heritage value categories "low", "medium", and "high" as a target (see Table and Figure 2b).

Table 1. Heritage value categories according to the classification by Stockholm's museum [45] and mapping of regional classifications to a common generic classification system.

| **Category** | **Description** (based on the city of Stockholm's classification [54]) | **Target category** |
|---|---|---|
| Blue | Highest class; buildings with exceptionally high cultural heritage value | High |
| Green | High cultural heritage value, particularly valuable from historical, culture-historical, environmental or artistic perspective | High |
| Yellow | Building has a positive meaning for city scape and/or has a certain culture-historical value | Medium |
| Grey | Does not have sufficiently high culture-historical value to be classified as one of the three previous categories | Low |
| Hatched | Buildings have not been classified, for example because they were too young | Low |

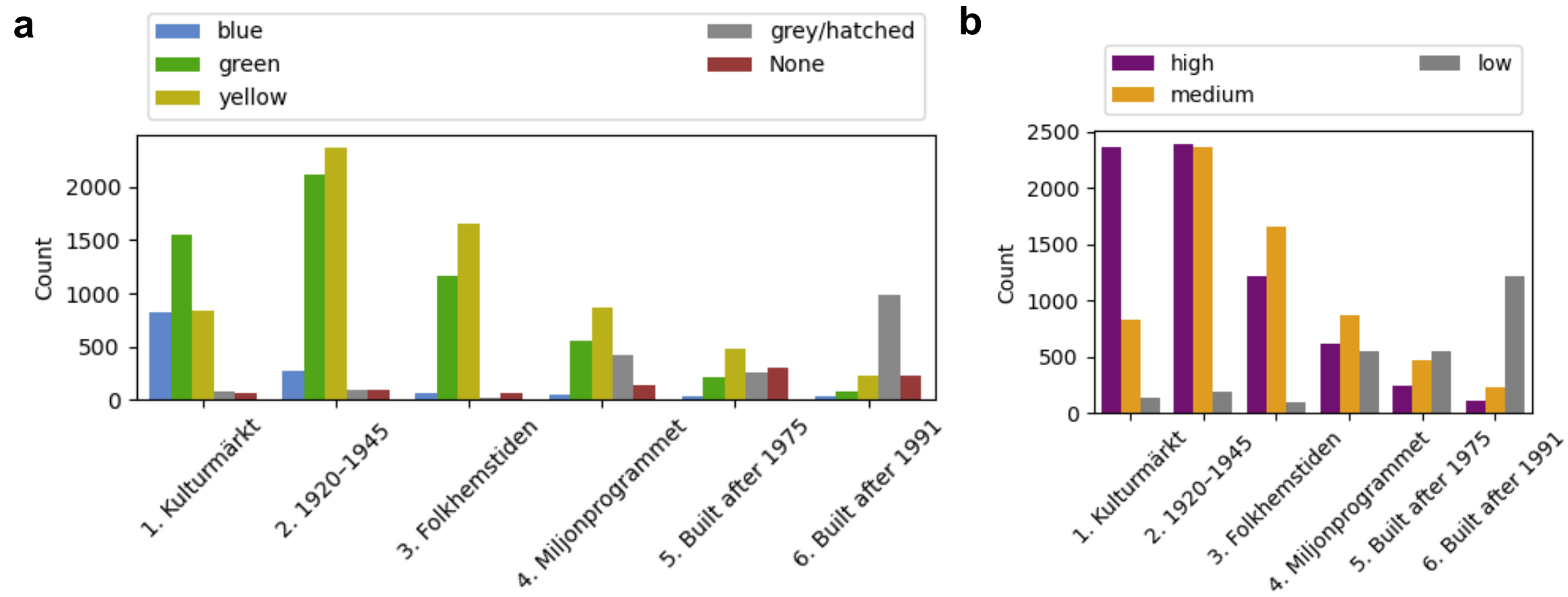

Fig. 2. Distribution of heritage categories in non-residential and multi-family buildings in Stockholm according to a) the official inventory of Stockholm's museum [54] and b) the simplified categorization used in this study. The x-axis shows construction year periods reflecting characteristic eras in the Swedish building construction history.

## 2.1 Retrieval of street-view imagery

The image data was retrieved from *Google Street View* [56] for development purposes using the same method employed by Sun et al. [13]. Currently, the usage of *Google Street View* for research and decision-making is ambiguous [57]. There is large potential for deriving features from street view imagery and *Google Street View* images are currently the most comprehensive repository. Alternative data sources are, for example, *Mapillary* [58].

The generation of images includes the following steps:

1. Extracting all energy-certified non-residential and multifamily buildings in Sweden from an internal building database developed at *RISE Research Institute of Sweden AB* as described in Johansson et al. [52] and Eriksson and Johansson [53].
2. Verifying street view image availability for the buildings using the *Google Street View Meta API* [59].
3. Calculating optimal camera points (see Fig3)
    a. Generalizing building footprint polygon to decrease number of vertices and separating into wall segments.
    b. Connecting the midpoint of each wall segment to the closest point on the road segment to create sightlines.
    c. Removing sightlines that are obstructed by another building, longer than 50 meters, or not perpendicular (tolerance -+3 degree).
    d. Selecting the shortest sightline if a wall had several.

e. Removing duplicate sightlines with similar angles (rounded to integers)
4. Retrieving actual camera locations close to the previously calculated optimal camera points through the *Google Street View Meta API* and recalculating and checking sightlines.
5. Determining the camera's pitch using the building's height from the energy performance certificate (number of floors * 3.0) along with the distance between the camera and the wall.
6. Calculating the zoom parameter based on the width of the wall and the distance between the wall and the photo point.
7. Downloading images from *Google Street View API* with the points location specified as longitude/latitude, heading, pitch, and zoom as input parameters.

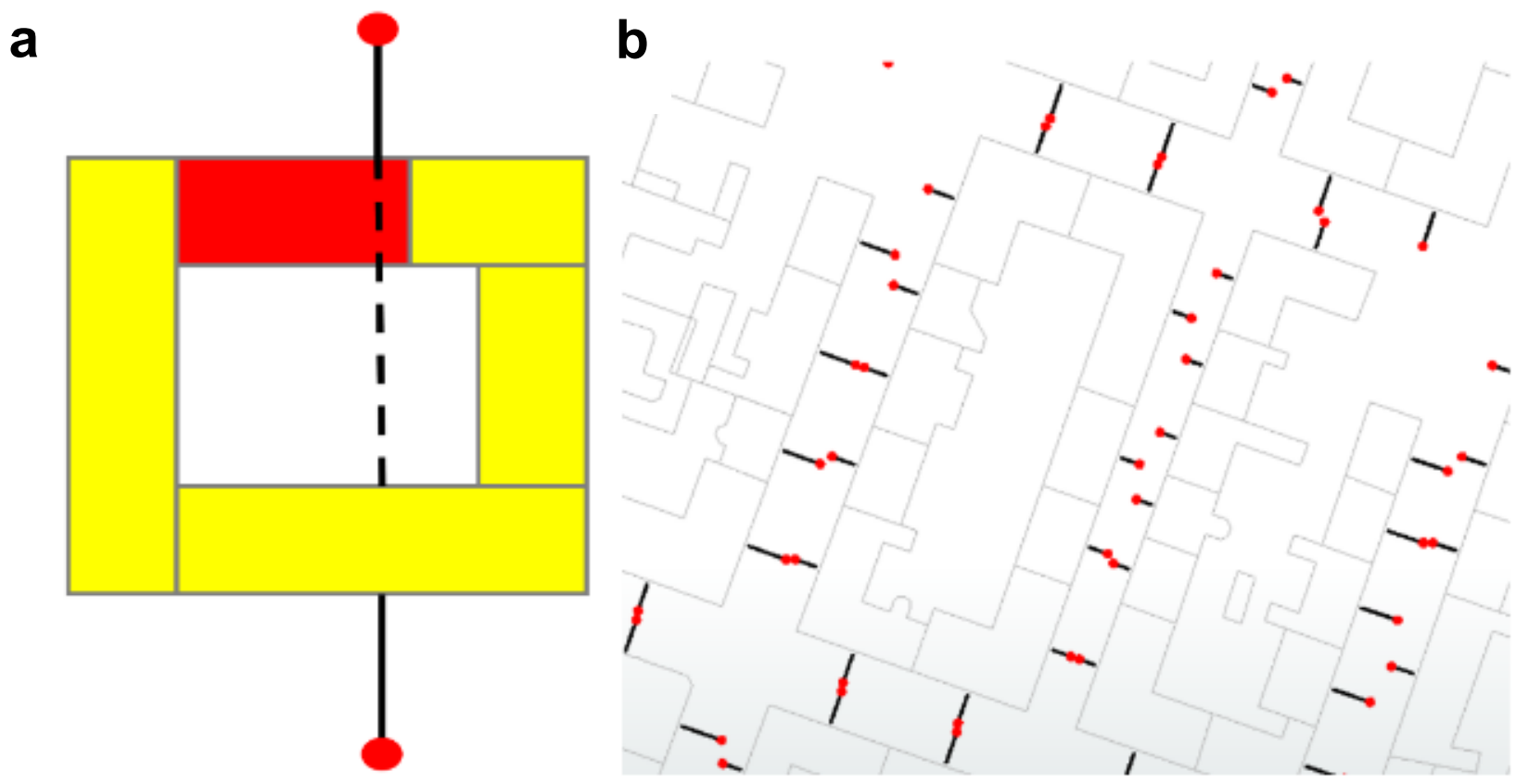

Fig. 3. Illustration of the selection of optimal camera points (red dots) for facade images of buildings (grey-outlined polygons) using perpendicular sightlines (black lines). a) Illustrates the removal of a sightline (dashed line) obstructed by other buildings (red polygon), b) shows an example of camera points (red dots) and sight lines (black lines) for rows of buildings (grey-outlined polygons).

## 2.2 Extraction of information using Large Language Models

There are several LLM available, and they evolve rapidly. For future replications it is advised to pre-test the quality of available LLM before applying the methods for the best results. When this analysis work was conducted OpenAI's GPT version gpt-4o-2024-08-06 had the highest benchmark results. The basic information sent to GPT in all cases was the facade image, which was first converted from binary data to ASCII characters using Python's base64 encoding algorithm [60]. The development of the prompt was an iterative process. The basis for the prompt were the authorities checklists of heritage values to be used during ocular assessments [61]. Heritage experts were part of the prompt design throughout the process. The final prompt used for the results presented in this article is provided as Supplementary Material S1.

Some of the information requested in the prompt requires more complex analysis [62], e.g., when asking which emotional reaction a building might evoke, while other information requests are more straight-forward [62], such as the number of balconies. Furthermore, prompts also follow a Template Pattern [63] and specify constraints, as suggested by Ekin [62]. Specifying constraints has the advantage of making the produced predictions more easily falsifiable. Accordingly, the final prompt specified valid response ranges (e.g., construction year 1000–2024) and categorical options for sub-questions, and the output format was set to JSON via the GPT API to standardize the responses.

An early iteration of the prompt was to attempt to extract a cultural heritage value category directly from GPT, however, this approach was discarded in favor of more detailed information extraction and postprocessing. GPT itself was included in the process, following the Question Refinement Pattern [63], i.e., asking GPT for help in formulating the question. Both in the question formulation phase and in the actual prompting, the Persona Pattern [63] was used, asking GPT to think as a heritage expert.

Even though the OpenAI guide [64] claims that temperature, a parameter for influencing the randomness of GPT [65], is not as relevant in GPT-4 as it was in previous versions, different temperatures were still evaluated to make sure settings produced consistent results, receiving the same results when running the same query for the same image multiple times, see Supplementary Material S2. For the final prompt the temperature was set as low as possible.

Other sources used in the design of the prompt included were literature studies on heritage value, facade characteristics, and architectural styles [32,66–68]. According to Beldjilali et al. [53], analysis of a building facade can be separated into an analysis of the facade itself, including for example, the style and construction period and the analysis of ornaments, consisting for example in the number and position of ornaments, repetitions and material. Facade ornaments were selected based on elements named in literature [31,32,66] and experience from manual classification that showed that some features, such as plaques, are generally not possible to detect in the images.

Furthermore, a workshop with heritage experts from the Swedish National Heritage Board was conducted on site in Visby on December 4th, 2024. The Swedish National Heritage Board were presented with facade images and asked to indicate what visual features contribute to or reduce the heritage values of the depicted buildings. Based on their input, the prompt was modified and additional features requested were added, in particular the number of floors and the share of window areas.

The final selection of features GPT was asked to extract, grouped by categories and alongside the allowed response values, is provided in Table S3 (in the Supplementary Material). The initial prompt engineering and overall methods development was carried out for a selection of buildings in Stockholm. The final prompt was applied to all multi-family-houses and non-residential buildings in Stockholm.

## 2.3 Postprocessing using machine learning

Postprocessing are the steps taken in assigning comparable heritage values based on the features generated by GPT. In the simplest case, this mapping can be done manually by basing the heritage category on a generated numeric heritage value on a continuous scale, by setting boundaries for classes and assigning low heritage values to the lowest heritage category and high heritage values to the highest category. However, setting this boundary manually is arbitrary and requires expert knowledge or systematic trial-and-error. Furthermore, more complex prompt responses require more sophisticated approaches.

Machine learning models were trained with the output of the LLM as input features and the heritage category as the target. The models that were tested are *XGBoost*, *KNeighborsClassifier*, *LogisticRegression* and *RandomForestClassifier* from the scikit-learn library [69]. Due to the different outputs of the prompts, the input features to the machine learning models varied. The target labels were retrieved by combining the predictions with the validation data. When training the machine learning models, a subset of 80% was used for training the model, 10% for validation during hyperparameter optimization and 10% for testing. The split between training, validation and test data was carried out in a stratified fashion based on the target variable, to make sure that both training and testing data contain all categories. The hyperparameter search spaces for the respective models are listed in Table 2. Sample weights were applied during training to mitigate the imbalanced training data.

Table 2. Hyperparameter search spaces for the models tested for heritage value category prediction.

| Model | Hyperparameter search space |
|---|---|
| RandomForestClassifier | "n_estimators": [50, 100, 200],<br>"max_depth": [None, 10, 20],<br>"min_samples_split": [2, 5], |
| LogisticRegression | "C": [0.1, 1.0, 10],<br>"penalty": ["l2"],<br>“solver": ["lbfgs", "liblinear"],<br>"max_iter": [1000] |
| KNeighborsClassifier | "n_neighbors": [3, 5, 10],<br>"weights": ["uniform", "distance"],<br>"p": [1, 2], |
| XGBoost | "objective": ["multi:softmax"],<br>"learning_rate": [0.01, 0.1, 0.2],<br>"max_depth": [3, 6, 10],<br>"n_estimators": [1000],<br>"subsample": [0.8, 1.0],<br>"early_stopping_rounds": [20],<br>"num_class": [len(y.unique())],<br>"enable_categorical": [True], |

## 2.4 Validation

Wang et al. [39] stress the importance of checking the stability and reliability of generative AI results within geospatial science. The validation approach presented in this study is comprised of two constituent parts: Firstly, an analysis of the features retrieved from the images by GPT and secondly, a validation of the final heritage value classification output.

Ground truth is available for validation for only some features such as construction year. The predicted construction years were compared with data from official statistics at the individual building level, and error metrics were calculated for named construction year epochs [53]. For the key parameter heritage value, we use Stockholm’s museum’s official inventory [54] as ground truth when assessing the consistency of GPT’s responses. However, this inventory does not extend beyond Stockholm. The strength of the correlation between categorical features and heritage categories was quantified using Cramér’s V, which is calculated according to Equation 1, with $X^2$ Pearson’s chi-squared statistic, $n$ the sample size, $k$ the number of columns and r the number of rows:

$$V = \sqrt{\frac{\frac{X^2}{n}}{\min(k-1, r-1)}} \tag{1}$$

In the second part of the validation, the results of the heritage value classification after postprocessing were compared against the heritage value classification data available for Stockholm [54]. As key performance indicators for comparing the various prompt-postprocessing combinations, precision and recall were calculated and based on these, the F1-score. Furthermore, the confusion matrix was retrieved, which provided information about the number of buildings classified correctly and incorrectly into the heritage categories.

# 3 Results

The workflow presented in Figure 1 is also reflected in this results section. First, the characteristics of the features are presented, showing the correlation of features with the heritage category and consistency across queries. Then, the importance of these features in the machine learning task are presented, and finally, the classification performance of the machine learning model is evaluated.

## 3.1 Feature characteristics

Construction year, though not in itself sufficient criterion for determining the heritage value of a building, is nevertheless an important factor. Figure 4 displays the prediction error of GPT's prediction compared with the construction year from the official registers. It can be observed that GPT has the highest errors for old buildings built before 1700, where it tends to predict that the buildings are younger than they are. This pattern continues for buildings up to approximately 1879, although the underestimation of the building's age is less pronounced here. Prediction errors are especially low in the period from 1880 to 1980, with average errors below 20 years. Errors increase again for younger buildings built after 1980. In this age class, GPT tends to underestimate the construction year and predicts the buildings to be older than they actually are.

When comparing the construction epochs, the predicted construction year more often fall into the epochs they actually were constructed in. GPT was more able to identify buildings constructed before 1920, with a precision and recall above 80% (see Figure 4d). Other epochs pose more problems, e.g., many buildings constructed between 1920 and 1945 were predicted to be younger and fall into the *Folkhemstiden* epoch. As old buildings constructed before 1920 are expected to be particularly relevant in terms of heritage value, it is promising that the majority of these buildings are correctly identified as belonging to this epoch.

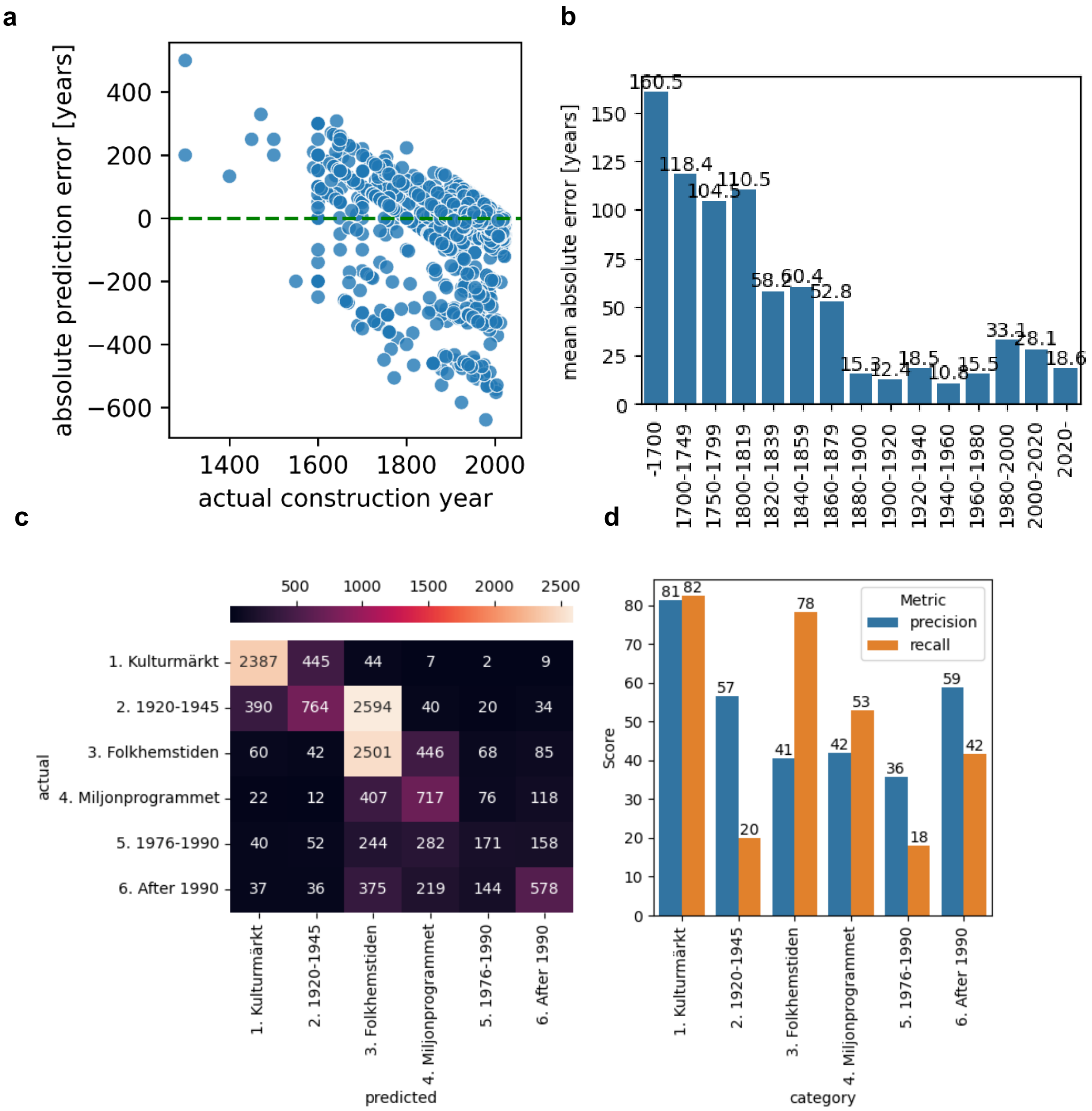

Fig. 4. Prediction error of the construction year from GPT compared to official registers; a) Individual prediction errors (positive = predicted younger, negative = predicted older); b) errors averaged over construction year periods; c) Confusion matrix of predicted vs. actual construction period (values show building counts); and d) precision and recall per construction period.

Apart from the construction year, GPT was asked for a range of other characteristics with different output types. Figure 5 displays the distribution of predicted values for the features rated on a scale from 1 to 100, split into the heritage value categories. The value distribution across the heritage categories is in many cases similar. For example, for state (renovation) and symmetry, the 25$^{th}$ and 75$^{th}$ percentile are identical in the boxplot. Furthermore, there are many outliers, indicating a large variability even within heritage value categories. For some features, however, the distributions differ more noticeably between the categories. This is the case for the features emotional reaction, aesthetic, rarity, and complexity, where values are higher for the high heritage value category than for medium and low. The larger difference between the heritage value categories are the features representative time, representative place and culture historical. This indicates that these features are the most promising for predicting the heritage category of a building.

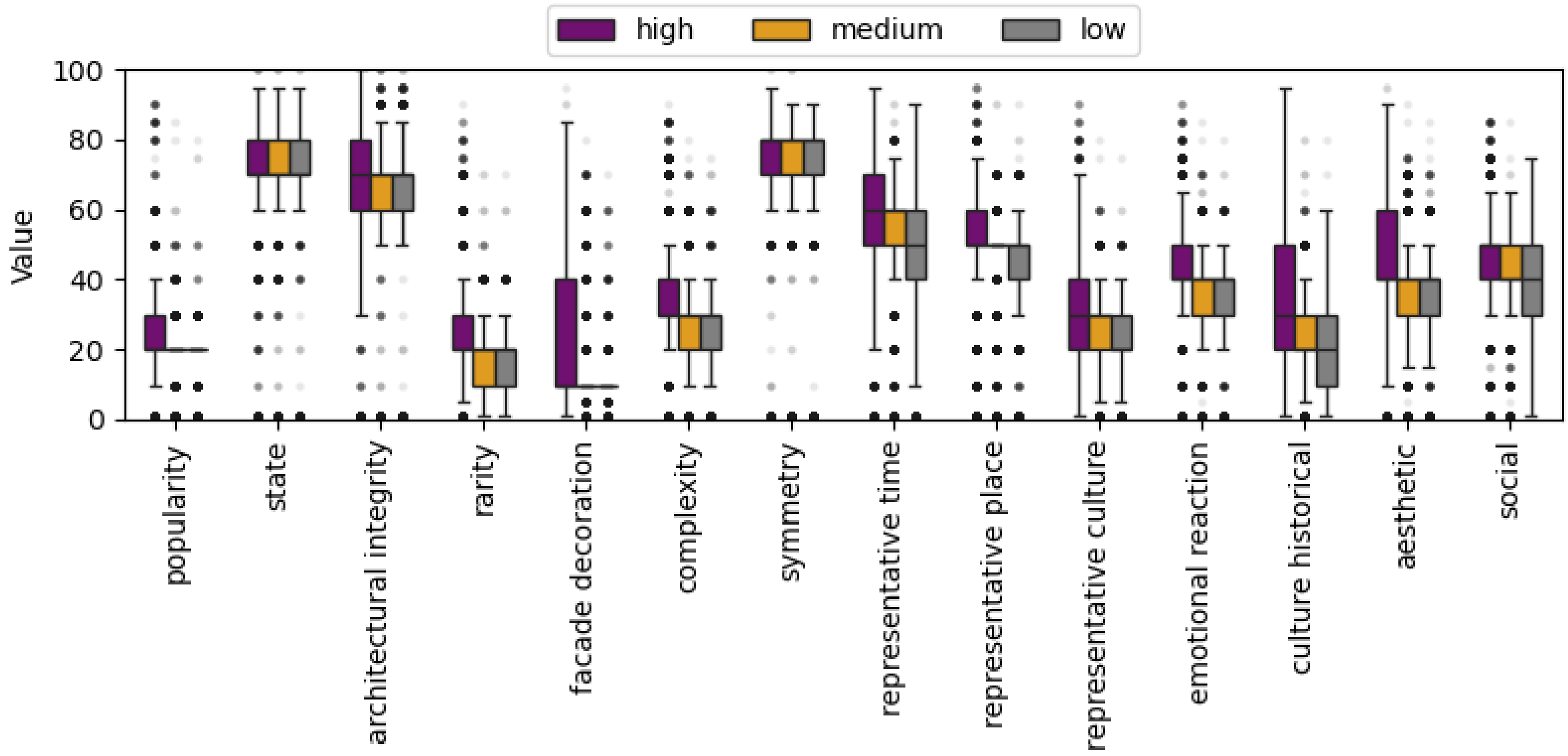

Fig. 5. Distribution of values for features rated on a scale by heritage category. Outliers are represented by transparent gray points, the darker the color, the higher the coincidence of occurrences for this value.

Figure 6 shows Cramér's V for categorical features. The highest value, i.e., the strongest association between the feature and the heritage value categories is observed for *style*, followed by the binary variable *door decorations* and the categorical variable *door shape*. These features are promising as predictors for the heritage value category of a building.

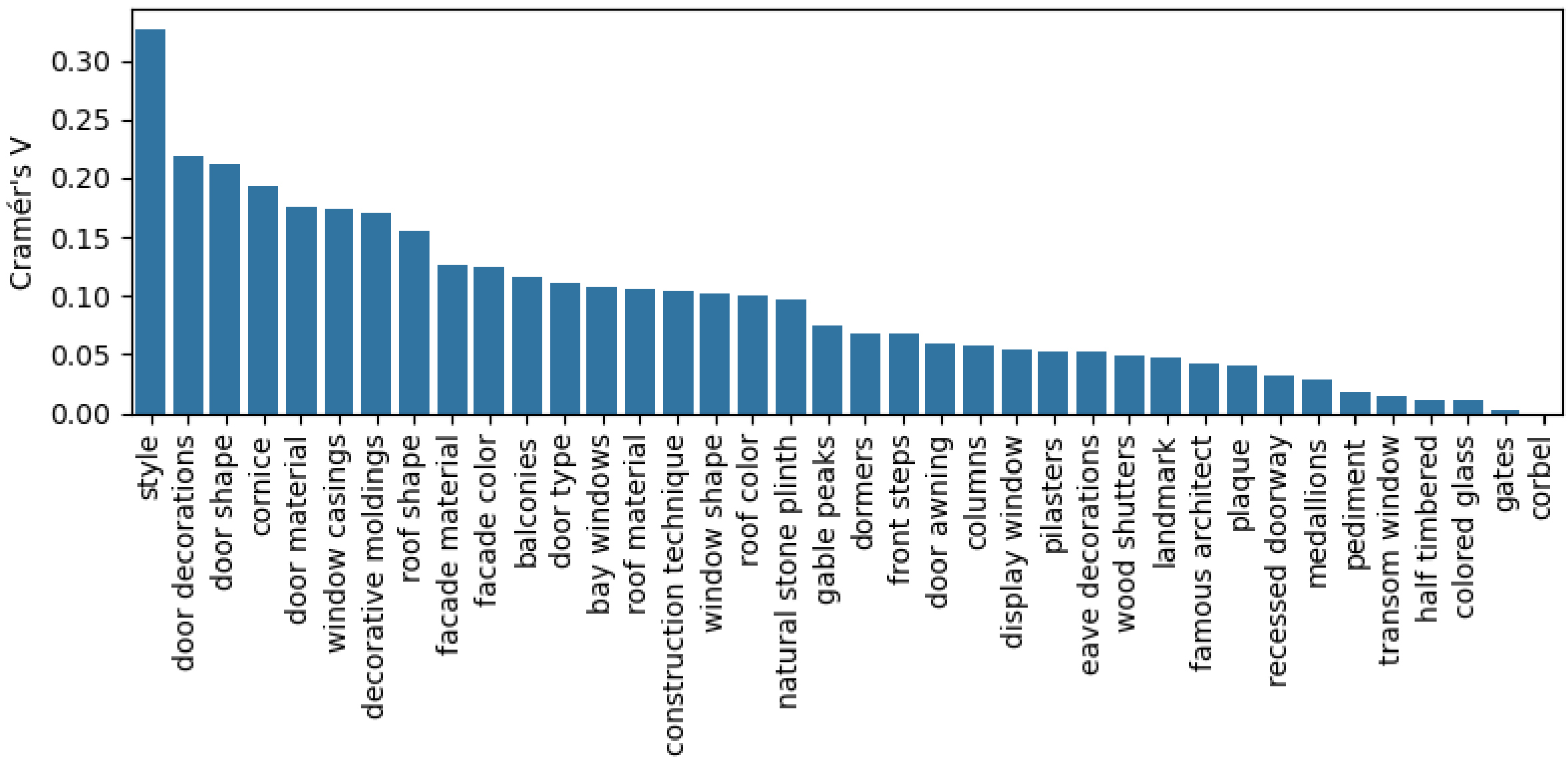

Fig. 6. Strength of association between categorical features and heritage categories, quantified by Cramér's V.

## 3.2 Classification performance

Figure 7 shows the macro F1-scores of the tested models using only the features retrieved from GPT (Figure 7b) and including the data on construction year, construction year period and building type available from the building database (Figure 7a). The best model in all scenarios is *XGBoost,* with a maximum F1-score of 0.71 when using all available data (F+cp+cy+t-XGB). Using only the data retrieved from GPT as features (F-XGB), the F1-score is 0.6. This shows that the model performance is higher when construction year and building type information is included in the model. The performance of the other models is lower, with the *RandomForest* model in second place when including the construction year and type information, and *LogisticRegression* in second place when using only data from GPT.

The F1-scores are compared against baseline models, namely a simple *stratified* assignment of heritage value categories for the case where no additional data is available and an *XGBoost* model trained on construction year, construction year period and building type alone, without features retrieved from GPT, for the case when this data is assumed to be available. The performance increase through the inclusion of data retrieved through GPT compared to using only the previously available building data is 0.05. Assuming that no building information apart from the GPT data is available, the performance of the model exceeds that of the *stratified* baseline model by 0.25.

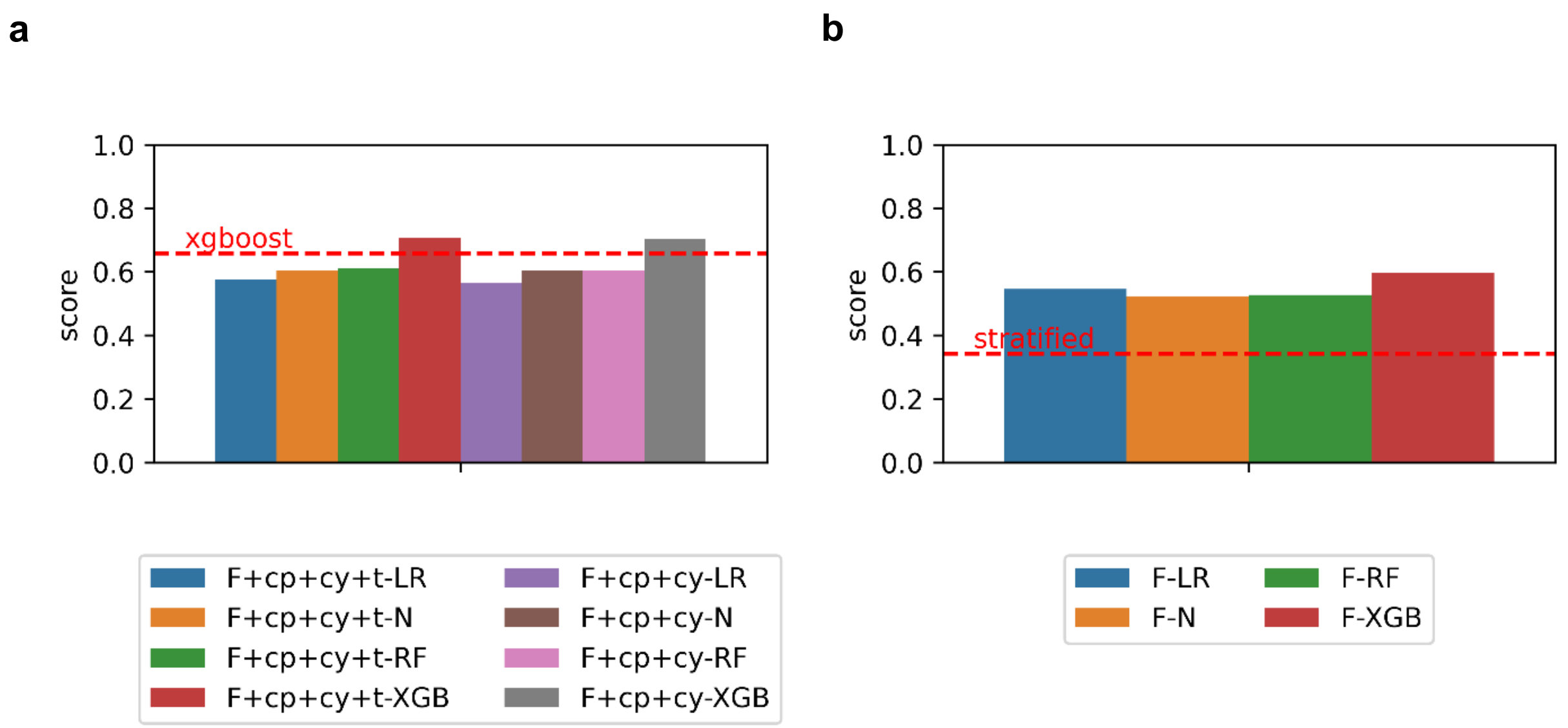

Fig. 7. Model Macro-F1 scores compared to baseline. a) Building-register + GPT features; (b) GPT extracted features only. Model naming: <features>-<classifier>. Features: F (GPT), cp (construction period), cy (construction year), t (building type). Classifiers: LR (Logistic Regression), N (K-Nearest Neighbors), RF (Random Forest), XGB (XGBoost)

Figure 8 shows the class-wise precision and recall for the model with the highest F1-score for both considered cases. Consistently with the F1-scores, precision and recall are higher overall when combining features retrieved from GPT with construction year and building type information. Furthermore, precision and recall are also more balanced. When training the model purely on the features retrieved through GPT, the recall performances for the highest and lowest heritage category decrease, compared to the first case. Combined with the higher recall of the medium heritage category, this suggests that the model is less able to distinguish between high/medium and low/medium, respectively, and tends to sort buildings into the medium category rather than in the highest and lowest. This pattern can also be observed in the confusion matrices.

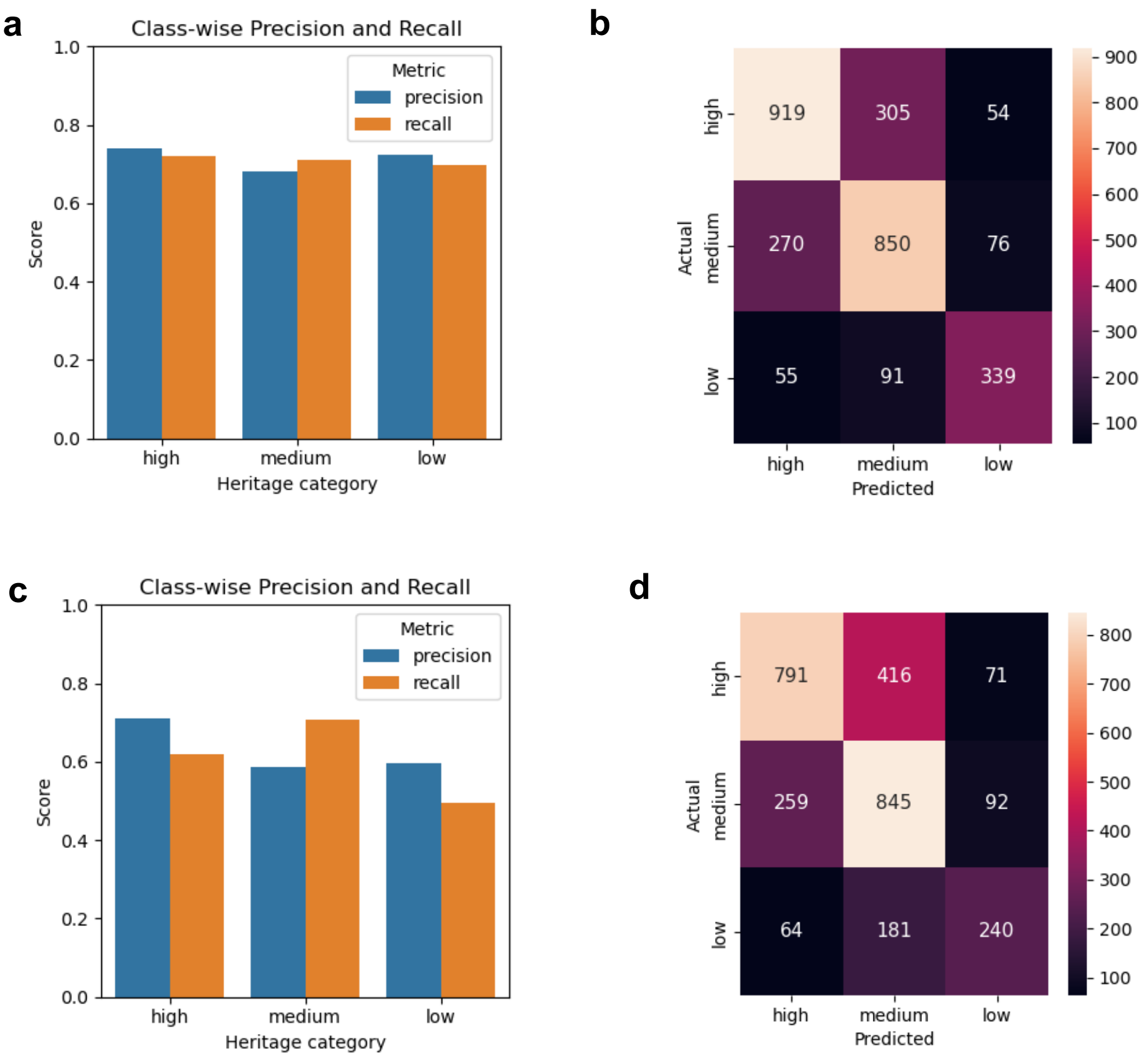

Fig. 8. Precision and recall as well as confusion matrixes for the model results. a) and b) show results when using all available data as features and c) and d) results when using only the features retrieved through GPT.

Evaluations of the feature importance of the *XGBoost* model found that style had the highest importance, followed by construction year information, see Figure 9. This is true in both cases, whether additional building data or only data retrieved from GPT was used. This shows that the construction year, even when predicted by GPT and with the inaccuracies presented above, is useful for determining the heritage value category. For the case that relied purely on GPT retrieved data, the culture-historical feature came in third. Followed by the GPT identified facade material and decoration. Interestingly, two of the three dimensions of heritage value described by the National Heritage Board [25], namely culture-historical and aesthetic values, appear within the top 10 features. Only the social dimension is not represented, which indicates that this dimension is most difficult to determine purely from facade images. Features appearing high up in the feature importance list are a result of the combination of the features being predictive of the heritage value category and GPT being able to non-randomly identify some characteristic GPT connects with the property asked for in the facade images. However, exactly what GPT identified cannot be evaluated, due to the black-box nature of the model and lack of detailed validation data.

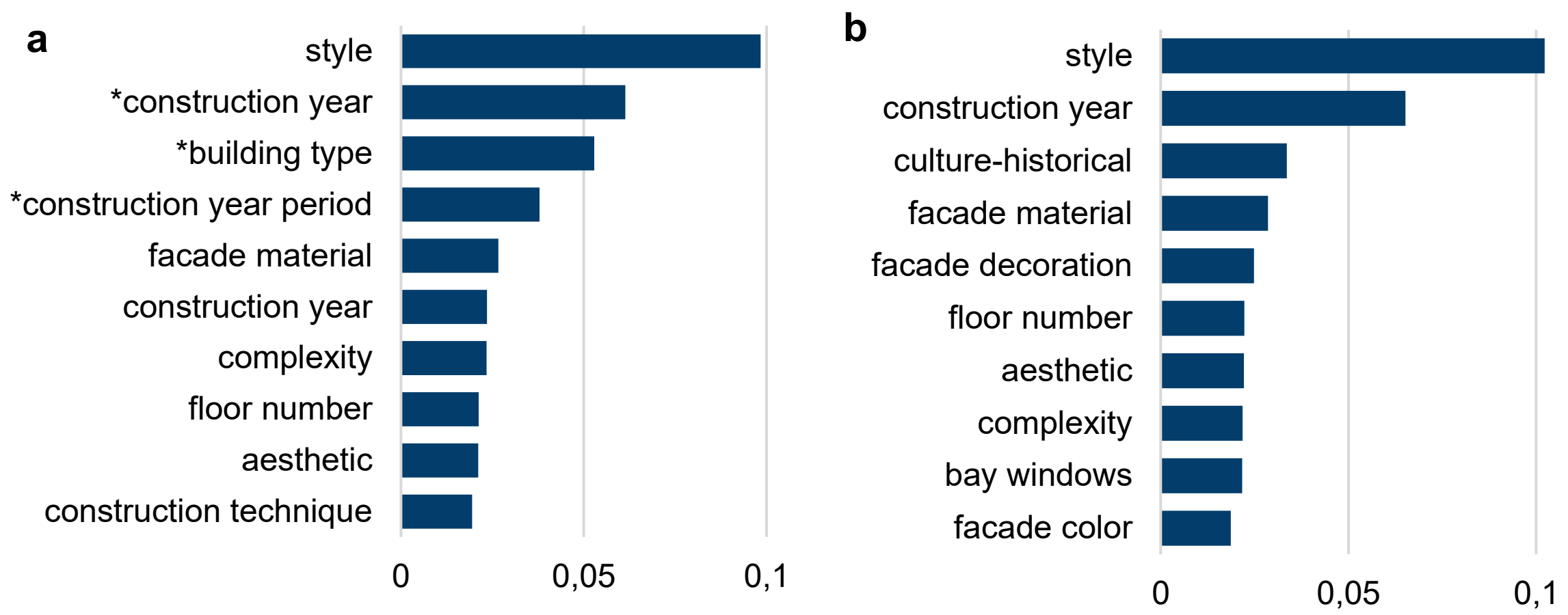

Fig. 9. Feature importance for the top 10 features: (a) model using all features, (b) model using only GPT-derived features. Asterisks (*) denote features obtained from building registers.

# 4 Discussion

The methods present promising results for the automatic assessment of features relating to heritage values of buildings based on street view images. The methods are particularly promising for large-scale data generation and analysis, which for example is required for the implementation of EPBD [7] in every European Member state. The approach saves both time and costs, as in-situ inventories require heritage experts to visit buildings in person [21].

Building heritage registers most often only include buildings which have a heritage value, buildings are often not registered as lacking heritage value, this makes training data imbalanced. Even though the Stockholm inventory includes a category for buildings without special heritage value, the proportion of buildings in this category is low. This is in part explained by the building stock of Stockholm being older than in other Swedish cities. However, it might also be caused by a general reluctance of heritage value experts to sort buildings into this category and thereby denying those buildings any heritage value. Imbalanced datasets can be mitigated by applying weights during training, random over- and under-sampling, depending on the specific case.

Construction year is central when assessing heritage value, also when human heritage experts assess building heritage. Whether or not to provide the construction year to GPT depends on the application case and the region considered. It should also be noted that there are different ways of including the building age in the workflow. The construction year could be included in the prompt, thus providing GPT with more information, but it can also be included in the postprocessing steps. The evaluation of the facade features will depend on the construction year; a building from the early 20th century will be characterized by the presence of ornaments and symmetry, whereas a modernist building could be characterized by the lack of ornaments and asymmetry. Other information that might improve the automatic assessment is neighborhood information. This might for example be provided through other input types such as aerial photography.

This study indicates that GPT had difficulties in predicting building age for newer buildings. Zeng et al. [14] had similar problems and surmised that this could be due to an underrepresentation of buildings of these classes in the testing data or due to less characteristic architectural features, it might also be caused by a tendency of GPT to respond with moderate instead of extreme values, avoiding the oldest and youngest classes. This could indicate risk-averse behavior, i.e., “prefer[ing] a deterministic outcome equal to the expectation of a risky outcome over that risky outcome” [70]. However, it should also be noted that even for heritage experts, determining the heritage value of more modern buildings is not straightforward. Dorset Council [68] state that it is difficult to determine whether or not post-war buildings should be listed, i.e., protected because of their heritage value. Furthermore, GPT is not trustworthy and was shown to outright ignore request. In one experiment, GPT was asked to guess the construction year of the building specifically based only on the image, even though GPT was given the correct age. In only 1 of 1,388 cases the estimated and the real construction year deviated.

The methods developed in this article can find buildings typical of those that have been considered of heritage value during the last 30 years. But since what we value changes with time, to keep records up to date or assess new types of buildings from a heritage value perspective, human intervention is necessary. Our findings agree with Chen et al. [71], GPT can support initial screening and data synthesis but cannot substitute for expert judgement, particularly when prioritizing and balancing heritage values that depend on contextual and tacit knowledge.

In the future, fine-tuning of the LLM could be an option. However, at the time of writing, GPT-4 does not support fine-tuning of vision [65]. Further limitations are that numbers provided by GPT may be approximates [65]. However, it is likely that the development of models will continue, providing better models with additional capabilities. Furthermore, a more detailed validation of the features extracted by GPT would be beneficial in order to gain a better understanding of its capabilities and limitations. However, the lack of ground truth data and the inherent complexity of validating subjective attributes such as “aesthetic value” make this challenging and are topics for future research.

In addition, testing the transferability of the model to other regions in Sweden and other countries would be valuable for deepening the understanding of cultural bias. For the Swedish case, there are problems in the comparability of existing inventories. Other regions in Sweden, such as Örebro and Halland, also have heritage value inventories. However, their categories differ from the ones used in Stockholm's inventory. Therefore, as a first step, a mapping between these categories would need to be conducted. Furthermore, the building stock of Stockholm is likely to differ from other cities, let alone rural areas in Sweden or regions in other countries. Therefore, new models would likely have to be trained again, including data from these regions.

This study raises several ethical questions. First, who has the right to determine which buildings have value and which do not. This question is not limited to the automatic assessment. Biases and blind spots are likely to exist both for heritage value experts and AI. Application of the methods requires broader discussion among relevant stakeholders and is a field for further evaluation. Another aspect is the energy usage of training and using AI. Both training AI models and using these models are energy intensive, with an AI-powered request having a higher electricity usage than standard Google searches, leading to concerns regarding future energy usage trends and impacts on sustainability [72,73]. However, this has to be set against the energy it would require carrying out the inventory by other means, which often include driving by car to examine buildings.

The Swedish authorities intend to use the predicted heritage values in various stages in the Energy Performance of Buildings Directive implementation during 2025–2026. The intention is also to provide energy experts with a handbook about energy efficiency measures in cultural heritage buildings along with the automated building specific assessment of heritage value. There will be a need to adapt the postprocessing steps for these separate tasks, as well as region specific conditions. It should also be noted that GPT is capable of producing very convincing but misleading descriptions, including assessments of architectural styles and facade elements of the buildings. Providing these to laypersons can be dangerous, as it may lead them to blindly trust the eloquently phrased assessments and neglect appropriate skepticism. Therefore, it is advisable to refrain from providing LLM-generated texts to cultural heritage laypersons, such as energy experts, and instead provide only categorical parameters, of which the validity is much easier for a layperson to assess.

# 5 Conclusion and outlook

This study shows the potential of using GPT for extracting information from facade images of buildings in order to determine aspects of their heritage value. Compared to using only previously known information about the construction year and building type of multi-family and commercial buildings in Stockholm, Sweden, integrating features retrieved from the images improved the classification performance of the *XGBoost* model by 5 percentage points. Compared to the best baseline model with no previously available building information, the *XGBoost* model trained on only features retrieved by GPT is 25 percentage points higher, underlining the potential, especially for regions where information on construction years is not available. The methods presented are promising because of the high workload and costs of manual inventories. However, it should not be forgotten that while automated methods can be an aid, heritage value experts should be included throughout the entire decision-making chain to mitigate errors.

The study offers a methodological contribution in an automated workflow that combines zero-shot analysis with GPT and uses the retrieved attributes as features for traditional machine learning classification. This contributes to a more nuanced understanding of GPT's capabilities and the applicability of machine learning approaches in the domain of heritage protection. Moreover, this approach has the potential to be applied in other fields, where the scarcity of labelled ground truth data and the consequent paucity of features hinder the training of machine learning models. The developed approach is of practical significance in large-scale building stock analyses, particularly when analyses require information that would otherwise not be available. One such example is the implementation of the EPBD, which stipulates that consideration must be given to buildings with high heritage value should be treated differently in energy retrofit planning. The theoretical contribution to the research field is to add the Stockholm case of building heritage value automated assessment.

An outlook and second step in the research will be to use the methods described in this article to assist Swedish authorities in the implementation of EPBD in all of Sweden. In that process the methods described will be used to facilitate expert validation for buildings beyond the Stockholm case. A future academic article [55] will include the various aspects of risks and mitigation strategies for authorities when using AI derived materials from the particular case of identification of heritage values for the EPBD implementation in Sweden. In that article, the transferability of the model to other regions within and outside of Sweden will be addressed. In addition, the methods could be adapted for other aspects relevant to the EPBD implementation and could be used to determine the building's usage or renovation state.

# Acknowledgements

We thank the Swedish National Heritage Board and members from Uppsala University's Conservation department for their input regarding the extraction of features from images, as well as for the valuable discussions regarding the applicability, potentials, and limitations of the methods.

This work was supported by the Helmholtz Association under the program "Energy System Design."

## Data availability statement

The Swedish register data are protected under confidentiality agreements. However, sharing of data is possible for research purposes in joint projects. The building image data is available from GoogleStreetView. A module enabling testing the methods and the source code is available at GitHub.

## Declaration of generative AI and AI-assisted technologies in the writing process

During the preparation of this work, the authors used DeepL for linguistic revisions. After using this tool, the authors reviewed and edited the content as needed and the authors take full responsibility for the content of the published article.

# Supplementary Material

## S1 Prompt

A prompt can be defined as “a set of instructions provided to an LLM that programs the LLM by customizing it and/or enhancing or refining its capabilities” [1]. Ekin [2] provides a list of techniques that should be observed when prompting GPT. Ekin states that instructions must be “clear and specific” and should include “explicit constraints”, as for example the length and format of the response. This goes hand in hand with verbally requesting a certain degree of output verbosity through the inclusion of keywords such as “briefly”. Furthermore, Ekin suggests including “context and examples” for improving results. Ekin differentiates between System 1 and System 2 questions, with the former asking for fact based, straight-forward responses and the latter requiring more complex analyses. White et al. [1] go beyond these suggestions by presenting systematical prompt patterns that can be employed for prompting LLMs. A prompt pattern consists of a name and classification, the intent and context, motivation, structure and key ideas and an example implementation. One of the Output Customization patterns is the Persona pattern, where the LLM is asked to take a certain role when replying to a prompt. Another pattern within this category is the Template pattern, which is used to specify the output structure. Within the context control category, the Context Manager pattern is used to provide contextual information to the LLM and specify what should and should not be considered. The Cognitive Verifier Pattern is based on the fact that LLMs are better able to answer questions if they are divided into sub questions [1]. Iterative testing and refining is one of the best practices mentioned by Ekin [2]. As stated by White et al. [1], the output quality is dependent on the prompt quality. Zeng et al. [3] found that GPT can use its knowledge to make predictions that go beyond the capabilities of traditional deep learning models.

In this study, the following prompt was used:

“As a cultural heritage value expert, assess the building in the image located at {{Adress}}. Valid values are provided in square brackets. When asked to rate on a scale, 1 represents minimal and 100 maximal value. Do not default to middle values unless they accurately reflect the assessment. When asked for a category, select exactly one category that fits best, unless explicitly asked to select multiple. Do not invent categories! If no evidence or external knowledge about the address is available, mark as false for boolean fields or assign a minimal score for numerical fields. Do not hallucinate! Return N/A option if it is impossible to answer the question, e.g., because the roof is not visible. Respond ONLY with a valid JSON object containing all the fields of the following dictionary. {construction_year: Estimated construction year. [1000-2024], famous_architect: Is the building associated with a famous architect? [true, false], landmark: Is the building a recognized landmark? [true, false], popularity: How popular is the building in the community? [1-100], state: How well maintained and undamaged is the building? [1-100], architectural_integrity: To what extent does the building appear to retain its original design, materials, and structural composition without visible modern alterations? [1-100], rarity: [1-100], style: [klassicism, romansk, gotik, renässans, barock, rokoko, nyklassicism, nygotik, nyrbarock, nyrenässans, nybarock, sekelskifte, nationalromantik, jugend, funktionalism, brutalism, high-tech, postmodernism, nyfunktionalism], construction_technique: [stolpverkshus, restimmerhus, resvirkeshus, plankhus, landshövdingehus, tegelhus, tjockhus, smalhus, lamellhus, punkthus, skivhus, burspråkshus], roof_shape: [flat, gabled, skillion, hipped, gambrel, pyramidal, crosspitched, sawtooth, cone, dome, onion, round, mansard, N/A], roof_material: [sheet metal, concrete, green, clay, copper, wood, straw, slate, bitumen, glass, asphalt, N/A], roof_color: [red, black, brown, green, grey, other, N/A], facade_material: [brick, concrete, wood, plaster, stone, metal, glass], facade_color: [red, yellow, white, blue, green, black, brown, grey, beige, other], facade_decoration: [1-100], window_area: What percentage of the total facade area is windows? [0-100], window_shape: [round, rectangular, rounded, square, N/A], window_number: [0-], window_avg_pane_number: Average number of panes per window. [1-], door_type: [single, double, portal, revolving, dutch, N/A], door_material: [metal, wood, glass, mixed, other, N/A], door_shape: [rectangular, arched, N/A], complexity: [1-100], symmetry: [1-100], floor_number: [1-], balcony_number: [0-], representative_time: How representative is the building for its construction time? [1-100], representative_place: How representative is the building for its location? [1-100], representative_culture: How representative is it for specific cultural, ethnical, religious, philosophical, or political expressions? [1-100], emotional_reaction: How positively people are likely to react to the building? [1-100], elements: Which elements are present? Choose ALL you can detect in the image: [balconies, bay_windows, dormers, gable_peaks, natural_stone_plinth, half_timbered, plaque, gates, colored_glass, wood_shutters, door_awning, front_steps, eave_decorations, window_casings, door_decorations, recessed_doorway, display_window, decorative_moldings, transom_window, pilasters, medallions, columns, cornice, tympanum, corbel, pediment], culture_historical: How does the building reflect historical events, specific time periods, and local activities? In what ways does it showcase different cultural, ethnic, religious, or philosophical influences? [1-100], aesthetic: Evaluate the building's overall architectural and artistic properties, e.g., facade composition, proportions, paintings, decorations, and ornaments. [1-100], social: Assess the attractiveness of the building and how positively it is perceived by different social groups. Consider which associations the building might evoke (e.g., beautiful, interesting, safe on the one hand, and ugly, boring, unsafe on the other) and how it serves the community. [1-100], visibility_score: How visible is the building in the image based on the following criteria: 1. Building is shown from the outside, 2. All floors and windows, as well as the roof, are clearly visible and not cut off. 3. It completely fills the image both vertically and horizontally. 4. No obstructions such as vehicles, vegetation, or scaffolding are covering significant parts of it. 5. Only one building is shown. [1-100]}”

# S2 Construction year consistency

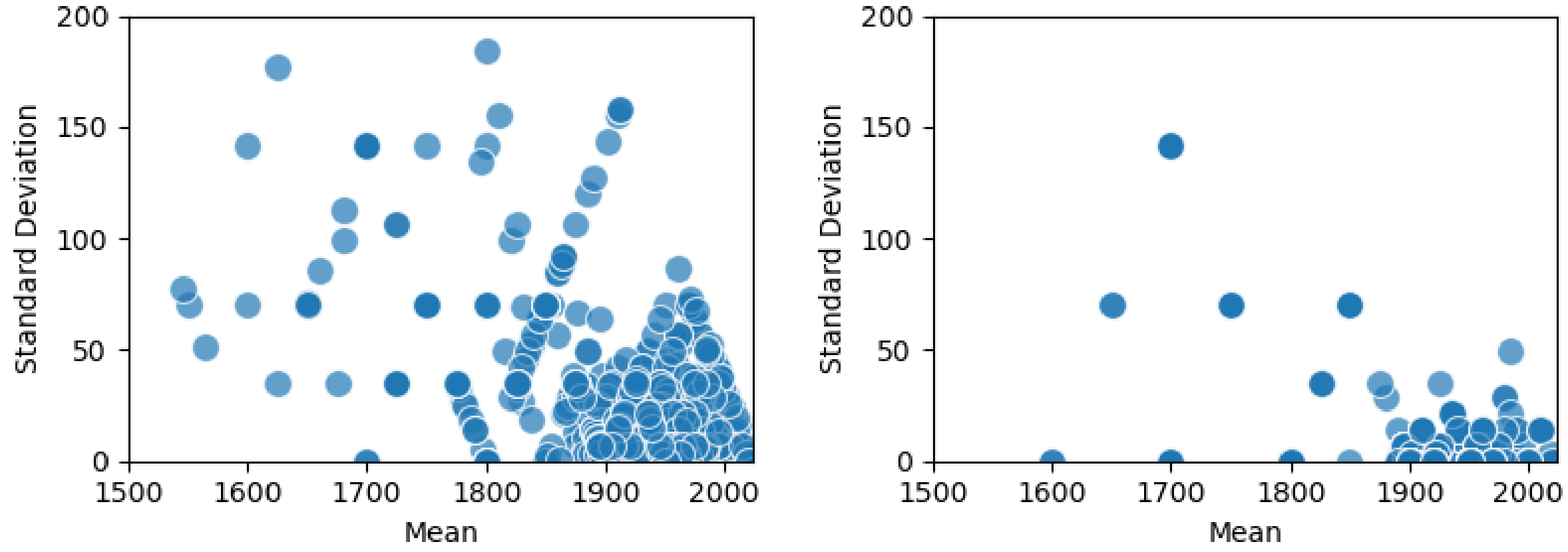

**Figure 1**. Standard deviation by mean construction year with a) high temperature (queries 15 and 16) and b) low temperature (queries 17 and 18); y axis limit set to 200 for display reasons.

**Table S3**. All features GPT was asked to extract from the street-level images.

| | **Feature** | **Values** |
|---|---|---|
| General | Construction year | 1000-2024 |
| | Famous architect | Yes/no |
| | Landmark | Yes/no |
| | Rarity | 1-100 |
| | Popularity | 1-100 |
| | Emotional reaction | 1-100 |
| | State | 1-100 |
| | Architectural integrity | 1-100 |
| | Style | klassicism, romansk, gotik, renässans, barock, rokoko, nyklassicism, nygotik, nyrbarock, nyrenässans, nybarock, sekelskifte, nationalromantik, jugend, funktionalism, brutalism, high-tech, postmodernism, nyfunktionalism |
| | Construction technique | stolpverkshus, restimmerhus, resvirkeshus, plankhus, landshövdingehus, tegelhus, tjockhus, smalhus, lamellhus, punkthus, skivhus, burspråkshus |
| | Number of storeys | 1- |
| Represent-ativeness | Time | 1-100 |
| | Place | 1-100 |
| | Culture | 1-100 |
| Forms, colors, materials | Roof shape | flat, gabled, skillion, hipped, gambrel, pyramidal, crosspitched, sawtooth, cone, dome, onion, round, mansard, N/A |
| | Roof material | sheet metal, concrete, green, clay, copper, wood, straw, slate, bitumen, glass, asphalt, N/A |
| | Roof color | red, black, brown, green, grey, other, N/A |
| | Facade material | brick, concrete, wood, plaster, stone, metal, glass |
| | Facade color | red, yellow, white, blue, green, black, brown, grey, beige, other |
| | Number of windows | Integer |
| | Share window area | 0-100 |
| | Window shape | round, rectangular, rounded, square, N/A |
| | Number of panes per window | Integer |
| | Door material | metal, wood, glass, mixed, other, N/A |
| | Door type | single, double, portal, revolving, dutch, N/A |
| | Door shape | rectangular, arched, N/A |
| Structure and ornaments | Complexity | 1-100 |
| | Symmetry | 1-100 |
| | Decorative elements | balconies, bay_windows, dormers, gable_peaks, natural_stone_plinth, half_timbered, plaque, gates, colored_glass, wood_shutters, door_awning, front_steps, eave_decorations, window_casings, door_decorations, recessed_doorway, display_window, decorative_moldings, transom_window, pilasters, medallions, columns, cornice, tympanum, corbel, pediment |
| | Decoration | 1-100 |
| | Number of balconies | Integer |
| Cultural heritage | Culture-historical | 1-100 |
| | Aesthetic | 1-100 |
| | Social | 1-100 |
| Other | Visibility | 1-100 |